\documentclass[10pt,twocolumn,letterpaper]{article}

\usepackage[pagenumbers]{wacv} 

\usepackage{graphicx}
\usepackage{amsmath}
\usepackage{amssymb}
\usepackage{booktabs}

\usepackage[pagebackref,breaklinks,colorlinks]{hyperref}

\usepackage[capitalize]{cleveref}
\crefname{section}{Sec.}{Secs.}
\Crefname{section}{Section}{Sections}
\Crefname{table}{Table}{Tables}
\crefname{table}{Tab.}{Tabs.}

\def\wacvPaperID{1300} 
\def\confName{WACV}
\def\confYear{2024}

\begin{document}

\title{Enhancing Monocular Height Estimation from Aerial \\Images with Street-view Images}

\author{Xiaomou Hou\\
The University of Tokyo\\
{\tt\small xiaomou.hou@ms.k.u-tokyo.ac.jp}
\and
Wanshui Gan\\
The University of Tokyo\\
{\tt\small gan@ms.k.u-tokyo.ac.jp}
\and
Naoto Yokoya\\
The University of Tokyo\\
{\tt\small yokoya@k.u-tokyo.ac.jp}
}
\maketitle

\begin{abstract}
Accurate height estimation from monocular aerial imagery presents a significant challenge due to its inherently ill-posed nature. This limitation is rooted in the absence of adequate geometric constraints available to the model when training with monocular imagery. Without additional geometric information to supplement the monocular image data, the model's ability to provide reliable estimations is compromised.

In this paper, we propose a method that enhances monocular height estimation by incorporating street-view images. Our insight is that street-view images provide a distinct viewing perspective and rich structural details of the scene, serving as geometric constraints to enhance the performance of monocular height estimation. Specifically, we aim to optimize an implicit 3D scene representation, density field, with geometry constraints from street-view images, thereby improving the accuracy and robustness of height estimation. Our experimental results demonstrate the effectiveness of our proposed method, outperforming the baseline and offering significant improvements in terms of accuracy and structural consistency.
\end{abstract}

\section{Introduction}
Height estimation from monocular imagery is a critical task in the remote sensing community. This task involves extracting the normalized Digital Surface Model (nDSM) or the absolute height data from a single aerial or satellite image. Understanding the geometry of a scene through this process has broad applications across various fields, such as urban planning~\cite{rs12172719}, biomass estimation~\cite{ten2019biomass}, and disaster management~\cite{KAKU2019417}. 

Techniques such as stereo vision~\cite{lemmens1988survey}\cite{flynn2016deepstereo} and Light Detection and Ranging (LiDAR)~\cite{PARK201976}\cite{park2018high} are commonly used for deriving height or depth information. Stereo imagery consists of pairs of images capturing the same scene from two cameras with slightly different viewpoints. Through a process known as stereo matching~\cite{lemmens1988survey}, corresponding pixels in the image pair can be identified. The disparity, or difference in position, between these corresponding pixels can then be used to estimate the depth or height of objects within the scene. LiDAR, on the other hand, directly measures the distance between the sensor and the object by emitting laser pulses and recording the time for the reflected pulse to return to compute the distance.
Despite their effectiveness in generating high-accuracy height maps, these methods come with their own limitations. For instance, Stereo vision requires multiple images and precise matching algorithms. As for LiDAR, it requires specialized equipment and has limitations in covering large areas, making it costly and challenging for tasks requiring frequent data updates over extensive regions.

In contrast, monocular imagery offers several advantages for height estimation. Monocular imagery requires only a single pass or sensor, representing a more cost-effective alternative, especially for large-scale tasks. Moreover, monocular imagery reduces the amount of data that needs to be processed and stored. Lastly, monocular imagery is often more readily available than its counterparts. For instance, many online mapping services, such as Google Maps~\cite{googlemaps} and Bing Maps~\cite{bingmaps}, provide extensive coverage of the earth through high-resolution satellite and aerial imagery. The widespread availability of such data ensures that we have a large amount of data for training and refining our models.

However, despite these advantages, estimating height from monocular imagery has some challenges. The primary challenge is that the geometry information is inherently lost when a 3D scene is projected onto a 2D image, particularly when the images are orthorectified, where the perspective effect is eliminated. Without a second viewpoint, it becomes difficult to recover the third dimension of the scene as contextual information, such as the actual size and relative distance of the objects, is not directly observable. This leads to the scale ambiguity problem~\cite{rs13122417}, which refers to the difficulty of determining the absolute size of objects from monocular imagery. Since an object imaged in a single image plane can have infinite correspondences in the real world, the height estimated from monocular imagery may not be reliable. 
In remote sensing specifically, this scale ambiguity problem amplifies due to the relatively long distance from the camera to the object. Given such remoteness, height differences between objects do not significantly alter their appearance in the imagery. For instance, two buildings of different heights might appear very similar in a monocular image due to the absence of noticeable perspective distortion. The remoteness, coupled with the absence of additional contextual information, makes accurate height estimation a particularly complex task in remote sensing imagery.

Considering these challenges, one potential solution is to integrate additional geometry cues from another source. Here, we propose to utilize street-view images, that is, images shot from the ground, for this purpose. Unlike aerial images, street-view images can provide a more detailed and close-up view of the objects, carrying explicit contextual clues like occlusion and perspective, which can be particularly useful for capturing the vertical structure information of ground objects such as buildings and trees. They contain visual cues related to geometry, such as the size of buildings relative to nearby objects or the perspective lines formed by building edges. This complementary information can help to disambiguate the height of structures and thus enhance the accuracy of height maps estimated from aerial images. Furthermore, street-view images are widely available and easily accessible today, thanks to platforms like Google Map Street View~\cite{googlestreetview}. This extensive availability makes them an ideal resource to leverage for augmenting monocular height estimation.

However, incorporating street-view images into a monocular height estimation model is not a straightforward task. 
The substantial difference in the viewing angles between the aerial and street-view images introduces a significant domain gap. This results in vastly different representations of the same scene, making it challenging for the model to reconcile the features extracted from these two types of images. Therefore, we need robust feature extraction and fusion strategies to bridge this domain gap. Next, the process of aligning street-view images with their corresponding aerial images is not always precise due to GPS errors, leading to a location drift problem. Lastly, the variation in scene content between the time of capturing aerial images and street-view images can also introduce extra noise to the model. For example, changes in vegetation, new constructions, or demolitions that occurred between the two capture times can lead to discrepancies between the two images, which can be challenging to handle.

In this work, we introduce the street-view constrained density field for enhancing height estimation. This method integrates geometry constraints from street-view images into an implicit 3D density field representation via volumetric rendering. Under supervision from both street-view images and ground truth height maps, we aim to optimize the density field as an intermediary of the scene geometry, thereby improving the accuracy of height predictions. Our experimental results indicate a significant improvement in the estimation accuracy and structural preservation in the height maps, demonstrating the effectiveness of our proposed methods. Our main contributions are summarized as follows.
\begin{itemize}
    \item We introduce the task of enhancing monocular height estimation with street-view images, which remains relatively unexplored in the remote sensing community.
    \item We extend the GeoNRW dataset~\cite{9406194} by integrating street-view images, enriching the existing resource for cross-view tasks.
    \item We propose a method that enhances monocular height estimation using a street-view constrained density field. Our proposed method outperforms the baseline significantly, demonstrating the effectiveness of street-view images and the flexibility of the density field in monocular height estimation.
\end{itemize}

The remainder of this paper is organized as follows. Sec.~\ref{sec2} provides a brief overview of works related to our study. Next, Sec.~\ref{sec3} elaborates on our proposals in detail. Following that, Sec.~\ref{sec4} introduces our experimental setups and conducts a comprehensive performance analysis of our methods. Finally, we conclude our study in Sec.~\ref{sec5}.

\section{Related Work}
\label{sec2}
\textbf{Monocular height estimation} is the task that derives the absolute height information from monocular remote sensing imagery, which is essential for understanding the 3D structure of extensive areas. Single overhead images lack vertical geometry information. 
To tackle this issue, Xie et al.~\cite{rs13152862} propose an indirect method that utilizes shadow analysis in remote sensing images to estimate building heights. Their method involves shadow extraction and regularization, as well as calculating the shadow length by combining the fish net line~\cite{liasis2016satellite} and the pauta criterion. This shadow length is then used in conjunction with the geometric relationship between the shadow and the building to estimate the building height.

Recently, the convolutional neural network (CNN)~\cite{schmidhuber2015deep}~\cite{long2015fully} has become the mainstream method for this task due to its capability of learning from a large amount of data to regress accurate height, making it a remedy to the ill-posed problem. Mou et al.~\cite{mou2018im2height} design a fully convolutional network that incorporates residual blocks~\cite{he2016deep} to regress height maps. Additionally, they employ a skip connection strategy to maintain the fine edge details in the prediction. 

Liu et al.~\cite{rs12172719} develop a network with skip connections and a postprocessing block. They propose a registration procedure to align optical data using LiDAR data. Additionally, the combined use of L1 norm, surface normal, and spatial gradient in their loss function improves accuracy and detail in the height estimation.

In contrast, Ghamisi et al.~\cite{8306501} reformulate the height estimation problem to an image-to-image translation problem, in which they employ the conditional Generative Adversarial Network (cGAN)~\cite{mirza2014conditional}\cite{goodfellow2020generative} to translate a single optical remote sensing image into the digital surface model (DSM). Following the pix2pix~\cite{isola2017image} framework, they not only use the objective function of cGAN to control authenticity but also add an L1 loss to enhance the accuracy.

\textbf{Monocular depth estimation} attracts more attention from the computer vision community compared to height estimation in the remote sensing community. Eigen et al.~\cite{eigen2014depth} make the first trial in monocular depth estimation using a CNN-based method. Their work presents a coarse-to-fine, multi-scale deep network designed for this task. The coarse network predicts overall depth, while the following network refines it. They also propose a scale-invariant loss function to mitigate scale ambiguity. 

Despite the reasonable results produced by optimizing a regression network, Fu et al.~\cite{fu2018deep} found that the convergence is slow due to the use of typical regression losses like L2 loss. To deal with the slow convergence problem, they propose a Depth Ordinal Regression Network (DORN) that incorporates the Spacing-increasing Discretization (SID) strategy and an ordinal regression training loss, converting depth estimation to a classification problem. 

Although CNNs demonstrate their promising performance in various computer vision tasks, they have limitations in expanding their receptive field to a larger scale. Transformers~\cite{vaswani2017attention}, on the other hand, consider the entire input sequence in computation, making them inherently capable of capturing global context. 
Ranftl et al.~\cite{ranftl2021vision} build upon the success of the Vision Transformer (ViT)~\cite{dosovitskiy2020image} and introduce a new architecture called Dense Prediction Transformers (DPT). This model shows a significant performance improvement in the general-purpose monocular depth estimation task compared to the best-performing CNN models. 

Since estimating either depth or height from monocular imagery is inherently ill-posed, Workman et al.~\cite{workman2021augmenting} argue that auxiliary data is needed to tackle this issue. They propose to fuse co-located overhead images into the street-view depth estimation network since they argue that overhead images provide the network with the scale of the scene, thereby mitigating the scale ambiguity. The results show that incorporating the geospatial context can significantly improve depth estimation accuracy. 

\textbf{Cross-view image synthesis} is the task that involves the interaction of overhead and street-view imagery, from which we can draw insights to bridge the two modalities. However, the complexity of this task escalates when the task involves synthesizing images with significant differences in viewpoints, as seen in the conversion between street-level and overhead images. 
Regmi et al.~\cite{Regmi_2018_CVPR} first address this task by proposing two cGAN-based architectures. The first architecture has a generator network with a two-head decoder for outputting segmentation maps as an auxiliary output to help guide the image generation process. The second architecture employs a sequence of two cGANs, generating better results than the first one. 

However, since no geometric constraints, such as height and depth, are fed into the cGANs, the synthesized images are likely to be distorted in structure. To fix this issue, Lu et al.~\cite{Lu_2020_CVPR} propose a framework that conducts satellite-to-ground synthesis in a geometrically meaningful way. They use street-view depth panoramas transformed from satellite depth maps to guide the BicycleGAN~\cite{zhu2017toward} to generate street-view images, as they argue that transformed depth panoramas can pose a strong inductive bias to the model, leading to diverse, structurally faithful results.

Qian et al.~\cite{qian2023sat2density} draw insights from the Neural Radiance Field (NeRF)~\cite{mildenhall2021nerf} and propose a cross-view image synthesis framework that utilizes volumetric neural rendering to learn the 3D representation of the scene without using real satellite height/depth maps. As a result, this work can generate distinct geometry from the input satellite image using the density field, and the flexibility of volumetric rendering enables end-to-end training through the whole pipeline.

\section{Method}\label{sec3}
\begin{figure*}
    \centering
    \includegraphics[width=1.0\linewidth]{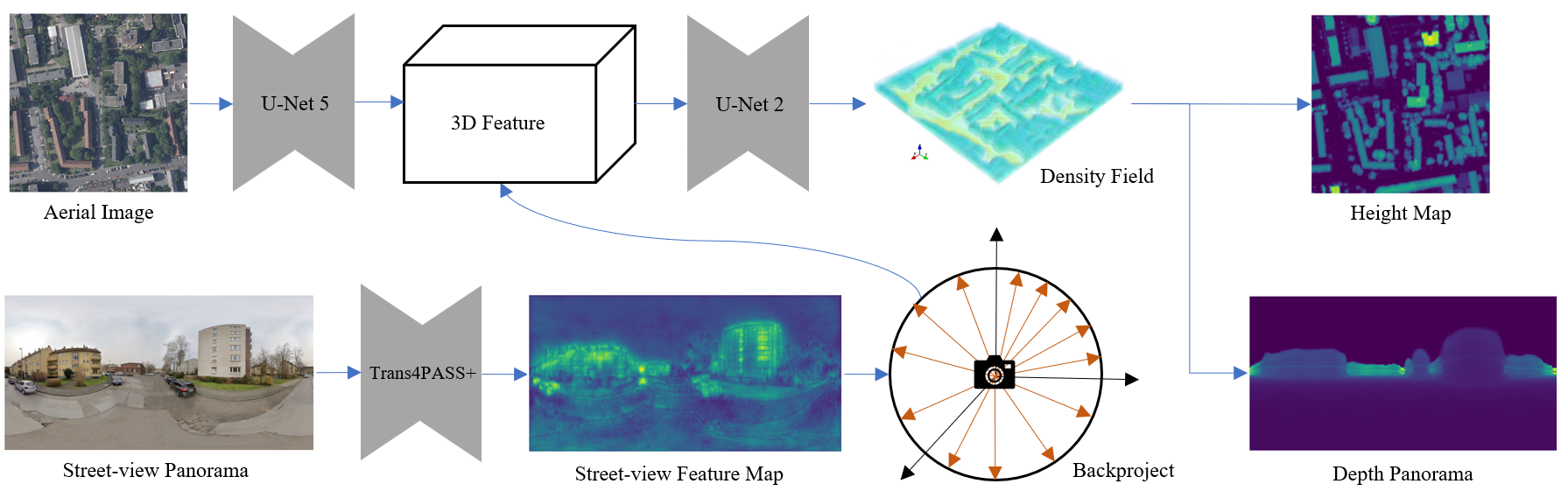}
    \caption{\textbf{Overview of our proposed method}. First, we extract image features for the aerial image and the street-view panorama, respectively. Next, these features are integrated into a unified 3D representation in 3D space. A shallow network then refines the fused feature to output the final density field. We conduct volumetric rendering on this density field to get both the height map and the street-view depth panorama, both of which are used to calculate the loss function for optimization.}
    \label{fig:overview}
\end{figure*}
In this section, we first give the mathematical definition of the problem of enhancing aerial image height estimation using street-view images. Next, we elaborate on our proposed method in detail. Our insight is leveraging street-view depth maps as constraints to the training process of the height estimation network, aiming to help mitigate the scale ambiguity by optimizing the 3D representation of the scene.

\subsection{Problem Definition}
The problem we address in this work involves using aerial and street-view imagery to estimate a height map of the corresponding area. Given an aerial RGB image $I_{a}$ and a co-located street-view RGB image $I_{s}$, the primary objective is to estimate a height map $H$. To achieve this, a neural network $f$, parameterized by $\theta$, is designed to output the height map $H$, such that $H = f(I_{a}, I_{s}; \theta)$. This neural network $f$ is trained with supervision from not only the ground truth height map $H_{gt}$ but also the street-view pseudo-depth map $D_{pseudo}$. 
The discrepancy between the estimated height map $H$ and $H_{gt}$ is quantified through a loss function $L_{H}(H, H_{gt})$. Next, we apply a transformation function $t$ to the estimated height map $H$, resulting in the depth map $D = t(H)$ that corresponds to $I_{s}$. We obtain the pseudo-depth map $D_{pseudo}$ from a pretrained model $g$, where $D_{pseudo}=g(I_{s})$. The discrepancy between $D$ and the pseudo-depth map $D_{pseudo}$ is quantified through another loss function $L_{D}(D, D_{pseudo})$.

Our goal is to minimizes the composite loss $L$, which is the weighted sum of the height estimation loss $L_{H}$ and the depth map discrepancy loss $L_{D}$, defined as follows:
\begin{align}
    L(H, D; \theta) = L_{H}(H, H_\text{gt}) + \alpha   L_{D}(D, D_\text{pseudo}),
    \label{eqt:total-loss}
\end{align}
where $\alpha$ is the weight that controls the contribution of $L_{D}$.

\subsection{Approach Overview}
The overall architecture of our proposed method is illustrated in Figure~\ref{fig:overview}. Given an aerial image and its co-located street-view panorama, we first apply two separate networks for feature extraction: for the aerial image, we use the U-Net~\cite{ronneberger2015u} to extract its 3D feature; for the street-view panorama, we leverage the Trans4PASS+~\cite{zhang2022bending}, which is a pretrained network dedicated for multi-scale feature extraction of panoramic images. The street-view panoramic feature map is then backprojected into 3D space, where it is fused with the 3D feature of the aerial image. Next, a shallow U-Net is applied to the fused feature to output the density field. Finally, we conduct volumetric rendering on the density field to obtain the height map and the depth panorama, which are used to calculate the loss function with the ground truth height map and the pseudo-depth map, respectively.

\subsection{Panoramic Ray-based Feature Fusion}
Since directly concatenating features from disparate modalities could potentially impair model performance, we propose a more reasonable fusion strategy, which is based on the imaging principle of panoramas, specifically equirectangular projection. Instead of direct concatenation, we backproject the panoramic feature map into 3D space and integrate it with the aerial image 3D feature map. Furthermore, drawing inspiration from ~\cite{cao2022monoscene}, our strategy involves backprojecting multi-scale 2D features along their respective panoramic rays and incorporating all potential 2D correlations into a unified 3D representation. The underlying hypothesis is that the resulting representation can guide the subsequent network from the aggregation of 2D features, thereby creating a bridge between the 2D and 3D feature spaces and allowing the subsequent network to self-discover relevant 3D features.

\begin{figure}
    \centering
    \includegraphics[width=1.0\linewidth]{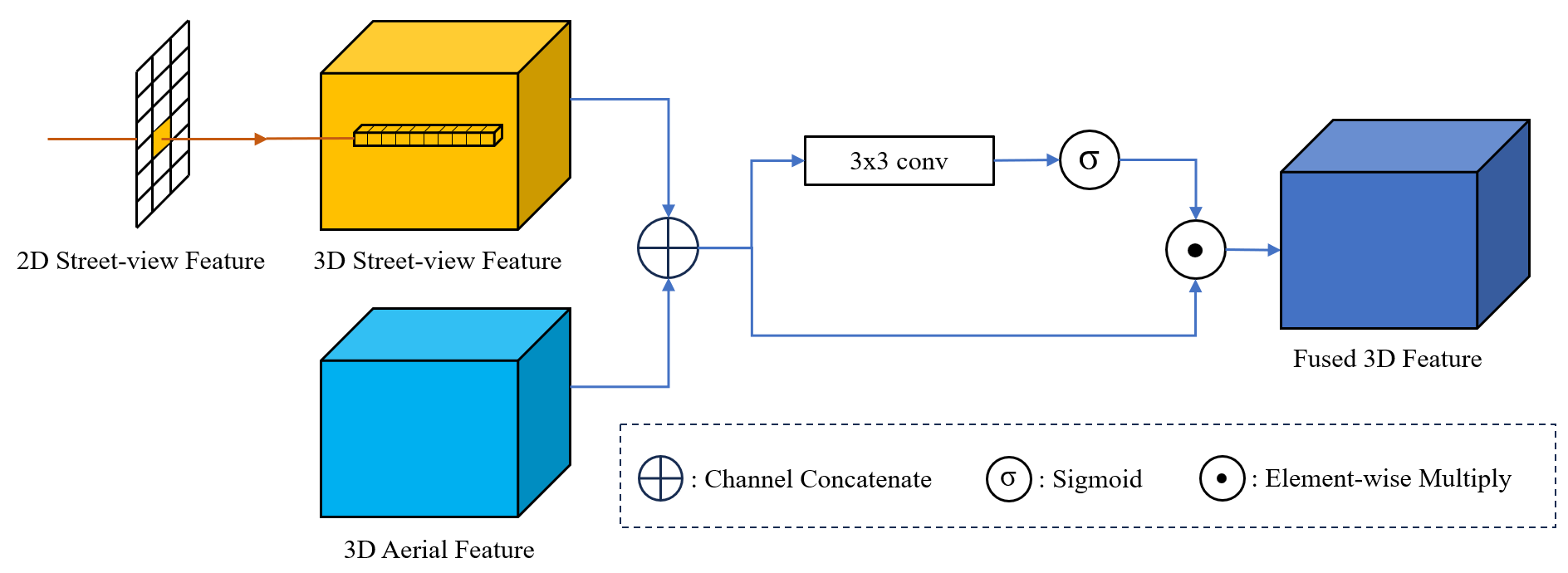}
    \caption{\textbf{The process of panoramic ray-based feature fusion}. This process starts with the backprojection of multi-scale 2D features along corresponding panoramic rays into 3D space. The subsequent phase involves applying the adaptive gated fusion to combine relevant features from different modalities.}
    \label{fig:fusion}
\end{figure}

As the street-view 3D feature is constructed, we fuse it with the aerial 3D feature using the adaptive gated fusion network~\cite{yoo20203d}, which selectively fuses the aerial 3D feature and the street-view 3D feature based on their relevance to the task. We apply channel-wise concatenation to the two 3D features and then use a 3x3 convolutional filter followed by a sigmoid function to get the attention map. The fusion is finalized by multiplying the attention map and the aerial 3D feature map. This process is depicted in Figure~\ref{fig:fusion}.

\subsection{Volumetric Rendering}
Inspired by ~\cite{mildenhall2021nerf} and ~\cite{qian2023sat2density}, we apply volumetric rendering to the density field output by the last U-Net to obtain the height map and the street-view depth map. This process involves setting a camera at the center of the density field, sampling the density values along multiple rays cast from the camera, and integrating these samples to generate the 2D depth map through the following rendering equation:
\begin{equation}
    \hat{d} = \sum_{i=1}^{S}T_i   (1-\exp(-\sigma_i   \delta_i))   d_i,
    \label{eqt:render}
\end{equation}
where $T_i=\exp(-\sum_{j=1}^{i-1}\sigma_j\delta_j)$, signifying the cumulative transmittance at sample point $i$. $\hat{d}$ represents the estimated depth value for each ray. $d_i$ is the distance between the camera location and the sampled point $i$. $S$ is the number of rays. $\sigma_i$ indicates the density at point $i$, and $\delta_i$ represents the step size.

The overhead height map and the street-view depth panorama are rendered in two different projection modes~\cite{shi2022geometry}, respectively.
For the height map, we apply a top-down parallel projection. The rendering process involves vertically casting rays downwards from a plane above the scene. We can compute the depth value of each ray based on Equation~\ref{eqt:render}. The overhead depth map is then inverted to get the height map. On the other hand, we adopt the equirectangular projection to render the street-view depth panorama. The direction vector of each sampling ray can be written as $\textbf{d} = (\cos\lambda   \sin\phi, -\cos\lambda   \cos\phi, \sin\lambda)$, where $\lambda$ represents the azimuth angle ranging in $[0,2\pi]$, and $\phi$ represents the elevation angle ranging in $[0,\pi]$.

\subsection{Pseudo-depth Map Generation}
With the street-view depth panoramas obtained, we aim to compute the loss function based on depth clues derived from the actual street-view color images. Since there are few depth estimation models with good generalizability for panoramic images, we aim to leverage depth supervision from perspective images. 
First, we extract the appropriate portions from the street-view panorama, creating a set of cutout perspective images heading $0^\circ$, $90^\circ$, $180^\circ$, and $270^\circ$. The field of view (FOV) is set to $90^\circ$, and pitch and roll keep $0^\circ$. Next, we utilize a pretrained depth estimation model, MiDaS~\cite{ranftl2020towards}, to generate the pseudo-depth maps for calculating the loss function. 

\subsection{Loss Function}
\begin{figure}
    \centering
    \includegraphics[width=1.0\linewidth]{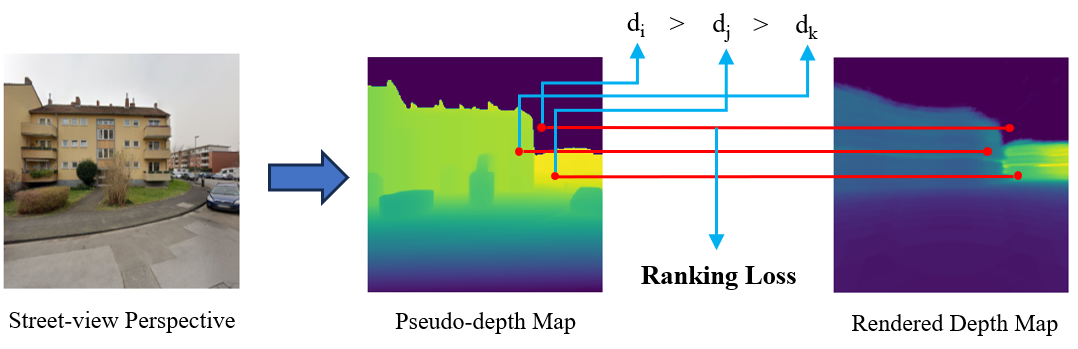}
    \caption{\textbf{The process of calculating the ranking loss}. First, we randomly sample point pairs in the pseudo-depth map. We identify the depth order of two points in the rendered depth map for each point pair. The depth order is compared against that in the pseudo-depth map to compute the loss.}
    \label{fig:ranking}
\end{figure}
We leverage supervision from ground truth height maps and street-view pseudo-depth maps. 
We use the scale-invariant loss~\cite{eigen2014depth} as our loss function for height maps, written as follows:
\begin{equation}
    L_{H}(H, H_\text{gt}) = \frac{1}{2N}\sum_{i=1}^{N}(\log y_i - \log \hat{y}_i + \zeta)^2,
    \label{eqt:scale-invariant-loss}
\end{equation}
where $\zeta = \frac{1}{N}\sum_{i=1}^{N}(\log \hat{y}_i - \log y_i)$, which is designed to penalize the scale deviation in the predicted height map. $y_i$ and $\hat{y}_i$ represent the $i$-th pixel value in the ground truth and predicted height maps, respectively. $N$ is the total number of pixels.

The loss for street-view depth maps is composed of 2 terms, written as $L_D=L_\text{rank}+L_\text{sky}$. 
The first term $L_\text{rank}$ is the ranking loss~\cite{chen2016single}\cite{lienen2021monocular}. This loss is beneficial for preserving the relative depth ordering and thus is suitable for leveraging supervision from pseudo-depth maps. The rationale behind this loss is that, given two points, A and B, if A is further away than B in the pseudo-depth map, the same relationship should hold true in the predicted depth map. The ranking loss is calculated based on pixel pairs, illustrated in Figure~\ref{fig:ranking}, and it can be written as follows:
\begin{equation}
    \begin{gathered}
        L_\text{rank}(D, R_{\text{pseudo}}, y) = \sum_{k=1}^{K}\Psi_k(D,i_k, j_k, r, y), \\
        \Psi_k = 
        \begin{cases}
          \log(1+\exp(-y_{i_k}+y_{j_k})), & r_k=+1 \\
          \log(1+\exp(y_{i_k}-y_{j_k})), & r_k=-1 \\
          (y_{i_k}-y_{j_k})^2, & r_k=0.
        \end{cases}
    \end{gathered}
    \label{eqt:ranking-loss}
\end{equation}
In the above equations, $R_{\text{pseudo}}$ is the rank matrix obtained from the pseudo-depth map, indicating the relative depth order for selected pixel pairs.
$K$ is the total number of query pairs.
$y_{i_k}$ and $y_{j_k}$ are the depth values at pixels $i_k$ and $j_k$ in the predicted depth map, respectively.
$r_k$ is the depth order indicator in the rank matrix $R_{\text{pseudo}}$: $+1$ if the depth at $i_k$ is less than at $j_k$, $-1$ if the depth at $i_k$ is greater than at $j_k$, and $0$ if they are equal.

In addition to the ranking loss, we add an extra sky-masked loss term $L_\text{sky}$ to the street-view depth map, as the sky masks can pose a strong inductive bias to the training process and thereby learn geometrically faithful density field, according to ~\cite{qian2023sat2density}. We obtain the sky masks by taking the same pretrained model~\cite{zhang2022bending} that we use in extracting features from street-view panoramas. With sky masks, $L_\text{sky}$ can be calculated as follows:
\begin{equation}
    L_{sky} = \frac{1}{N}(\sum_{r\in R} \lvert \hat{O}(r)-1 \rvert + \sum_{r\in R'}\lvert \hat{O}(r)\rvert),
    \label{eqt:sky-masked-loss}
\end{equation}
where $\hat{O}=\sum_{i=1}^{S}T_i (1-\exp(-\sigma_i   \delta_i))$, which is the street-view opacity derived from the density field. $R$ and $R'$ represent the rays pointing to the non-sky and sky regions, respectively. $N$ is the total number of rays. The first term of $L_{sky}$ penalizes the estimated opacity deviating from $1$ (completely opaque) for the non-sky region, and the second term penalizes non-zero opacity for the sky region.

\section{Experiments}\label{sec4}
\subsection{Dataset}
Our dataset is built upon the GeoNRW dataset~\cite{9406194}, which is acquired in Germany, consisting of orthorectified aerial photographs, LiDAR-derived digital surface models (DSMs), and segmentation maps with $10$ classes, with a resolution of $1$m and the size of $1000\times1000$ pixels. Out of its total $7,783$ image triplets, we use $881$ triplets acquired in Cologne and Dortmund, Germany. In addition, since the DSM represents the sum of the height of the ground and the height of the above-ground objects, we convert DSM to the normalized digital surface model (nDSM), which represents the absolute height of the above-ground objects. This process is done with the help of ~\cite{dsm2dtm}.

Our street-view images are obtained through the Google Map Street View Static API~\cite{googlestreetview}. First, we apply a uniform grid with a vertical and horizontal spacing of $100$ pixels on each aerial image tile, resulting in an array of grid cells across the aerial image tile. For each grid cell, we identify the center point and retrieve the available street-view image closest to this point.
After acquiring these street-view images, we return to aerial image tiles and nDSMs and crop out corresponding patches centered at the exact location of retrieved street-view images, ensuring a precise correspondence between each street-view image and its matching aerial image patch. Each patch is sized at $256\times256$, covering the area visible in the associated street-view images. As a result, we obtain $11,441$ data groups. The train-validation-test split is set to 8:1:1.

\subsection{Evaluation Metrics}
In our experiments, we utilize the following three evaluation metrics:
\begin{itemize}
    \item $\text{Mean Absolute Error (MAE):}\frac{1}{N}\sum_i^n\lvert y_i-\hat{y}_i\rvert$, where $N$ is the total number of samples, $y_{i}$ denotes the ground truth height value, and $\hat{y}_{i}$ denotes the predicted height value. The MAE measures the average absolute difference between the ground truth and the predicted height values generated by our models.
    \item $\text{Root Mean Square Error (RMSE):}\sqrt{\frac{1}{N}\sum_i^n (y_{i} - \hat{y}_{i})^2}$, where the notation keeps the same as that of the MAE. The RMSE emphasizes larger errors by squaring the differences before taking the average.
    \item Structural Similarity Index Measure (SSIM)~\cite{wang2004image}: $\frac{(2\mu_y\mu_{\hat{y}} + c_1)(2\sigma_{y\hat{y}} + c_2)}{(\mu_y^2 + \mu_{\hat{y}}^2 + c_1)(\sigma_y^2 + \sigma_{\hat{y}}^2 + c_2)}$, where $y$ and $\hat{y}$ are patches of the actual and predicted height maps, respectively. $\mu_y$ and $\mu_{\hat{y}}$ are the averages of $y$ and $\hat{y}$, respectively, and $\sigma_{y\hat{y}}$ is the covariance of $y$ and $\hat{y}$. $c_1$ and $c_2$ are two variables to stabilize the division with a weak denominator. The SSIM measures perceptual and structural similarity between our model's height estimations and the ground truth.
\end{itemize}

\subsection{Implementation Details}
In our proposed method, we employ a 5-layer U-Net for outputting the aerial image 3D feature with a size of $256\times256\times64$. The street-view panorama is processed through ~\cite{zhang2022bending}, resulting in a 2D feature map, which is then transformed into a 3D feature for fusion. The fused 3D feature undergoes refinement by another 2-layer U-Net to yield the final density field with a size of $256\times256\times64$. For our loss function, we set the weights of both $L_{H}$ and $L_{D}$ to $1.0$. In addition, to calculate the ranking loss in Equation~\ref{eqt:ranking-loss}, we randomly sample $2,048$ point pairs of each street-view image and constrain the distance between two points in each point pair to be $10$ to $30$ pixels.

Our implementation is based on PyTorch~\cite{NEURIPS2019_9015}. We set the number of training epochs to $200$ and the batch size to $8$. We use the Adam optimizer~\cite{kingma2014adam} with a learning rate of $0.0001$, and the parameters $\beta_1$ and $\beta_2$ are set to $0.9$ and $0.999$, respectively. Additionally, we employ the step learning rate decay strategy, where the learning rate is multiplied by $0.1$ every $100$ epochs. We train our models with two NVIDIA 2080 Ti GPUs.

\subsection{Results}
\begin{table*}[ht]
  \centering
  \caption{\textbf{Quantitative evaluation} ($\downarrow:$ lower is better; $\uparrow:$ higher is better; \textbf{bold} indicates the best). \textbf{Baseline} denotes the standard method of estimating height maps. \textbf{Density} applies only the density field; \textbf{Density + Street Loss} adds the street-view loss; \textbf{Density + Fusion} incorporates feature fusion; and \textbf{Density + Fusion + Street Loss} combines all the three techniques.}
    \setlength{\tabcolsep}{7mm}{
    \begin{tabular}{lccc}
    \toprule[1pt]
    \multicolumn{1}{c}{$\text{Methods}$} & \multicolumn{1}{c}{$\text{MAE}\downarrow$} & \multicolumn{1}{c}{$\text{RMSE}\downarrow$} & \multicolumn{1}{c}{$\text{SSIM}\uparrow$}\\
    \midrule[.5pt]
    Baseline & $1.2550_{\pm0.0041}$  & $1.9656_{\pm0.0029}$ & $0.7151_{\pm0.0056}$ \\
    Density    & $1.2306_{\pm 0.0031}$  & $1.9316_{\pm 0.0039}$ & $ 0.7242_{\pm 0.0019}$ \\
    Density + Street Loss & $1.2198_{\pm 0.0089}$  & $1.9081_{\pm 0.0068}$ & $0.7076_{\pm 0.0078}$ \\
    Density + Fusion & $1.1823_{\pm 0.0041}$  & $1.8550_{\pm 0.0051} $ & $\textbf{0.7246}_{\pm 0.0026} $ \\
    Density + Fusion + Street Loss & $\textbf{1.1750}_{\pm 0.0027}$  & $\textbf{1.8426}_{\pm 0.0034}$ & $0.7178_{\pm 0.0039}$ \\
    \bottomrule[1pt]
    \end{tabular}}
  \label{tab:quantitative}
\end{table*}

\begin{figure*}[htp]
    \centering
    \includegraphics[width=0.93\linewidth]{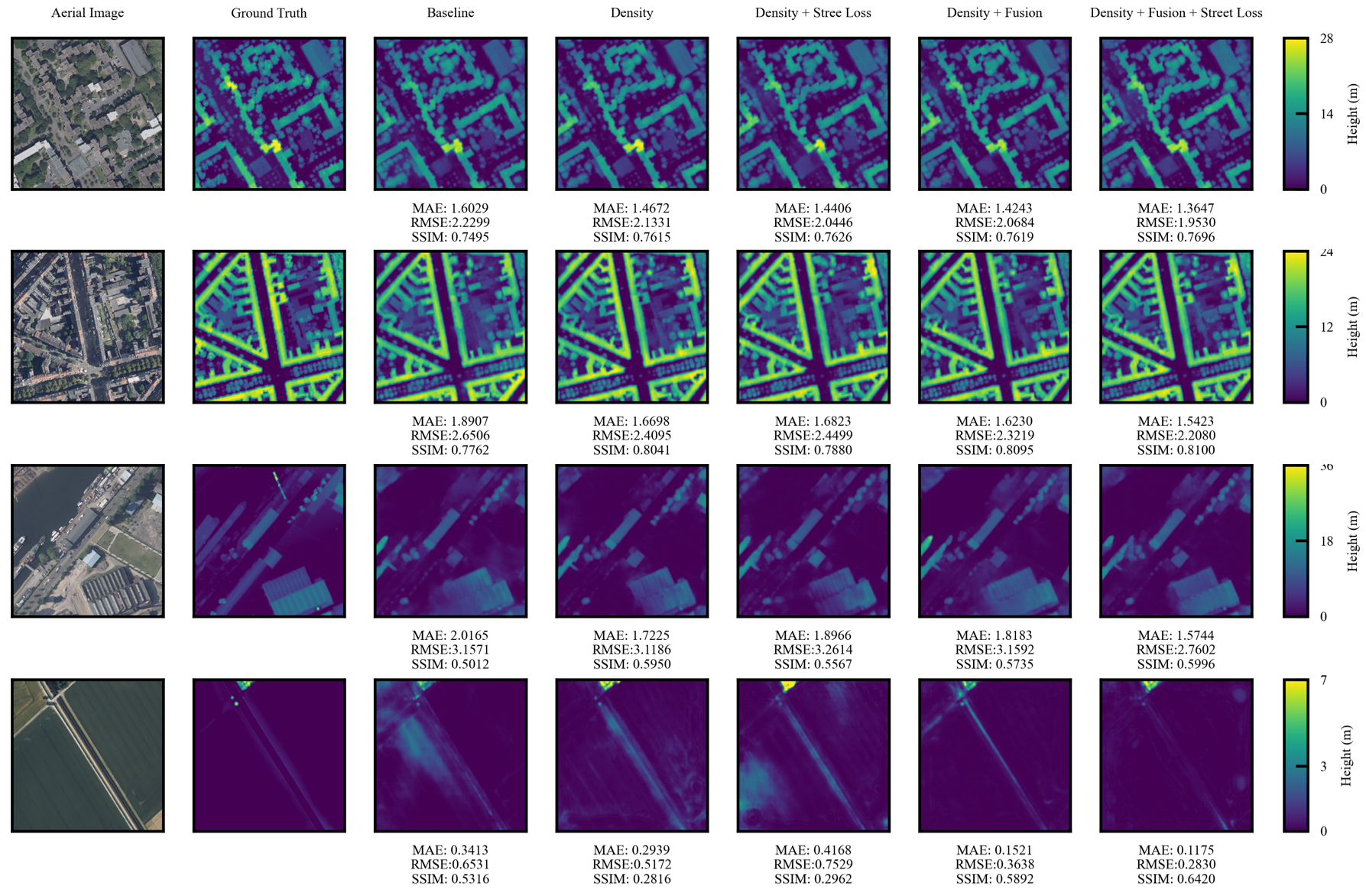}
    \caption{\textbf{Qualitative evaluation}. \textbf{Baseline} denotes the standard method of estimating height maps. \textbf{Density} applies only the density field; \textbf{Density + Street Loss} adds the street-view loss; \textbf{Density + Fusion} incorporates feature fusion; and \textbf{Density + Fusion + Street Loss} combines all the three techniques.}
    \label{fig:qualitative}
\end{figure*}

We conduct comprehensive performance assessments of our results, including quantitative and qualitative evaluations. Our baseline model employs a 5-layer U-Net, taking only aerial images as input. It regresses a $256\times256\times1$ height map with only the ground truth height map as supervision. The loss function is the scale-invariant loss in Equation~\ref{eqt:scale-invariant-loss}.

\subsubsection{Quantitative Evaluation}
From the quantitative results in Table~\ref{tab:quantitative}, we notice that solely using the density field offers a slight performance improvement compared to the baseline. This can be attributed to the density field's ability to harness multiple channels for reconstruction, in contrast to the single-channel regression in the baseline. Next, incorporating the street-view loss enhances MAE and RMSE results but sees a dip in SSIM, hinting at potential noise from the street-view data affecting the structural consistency. On the other hand, when we merge street-view and aerial image features, there is a notable uplift in performance across all metrics. This fusion strategy offers a rich blend of local street details and aerial context, leading to precise height predictions. Lastly, combining fusion and the street-view loss yields the best MAE and RMSE, but there is a decrease in SSIM. This compromise can be attributed to the inherent challenges of fitting two different modalities: the domain gap between street-view and aerial images and the inevitable noise in the street-view pseudo-depth map. This noise can perturb the output structural similarity, reflecting the balance our model must maintain when jointly optimizing for both data sources.

\subsubsection{Qualitative Evaluation}
From Figure~\ref {fig:qualitative}, we can observe that, with ground geometry information, our model can holistically mitigate the scale ambiguity, reflected by the dropped MAE and RMSE. Furthermore, our model demonstrates its generalizability in both areas with high building density and areas with flat terrain, and its effect is more noticeable in flat areas compared to dense areas. 

Consequently, our proposed method outperforms the baseline in all aspects, indicating the benefits of leveraging street-view data for aerial height estimation. These qualitative results further corroborate the findings from our quantitative analysis, reaffirming the effectiveness of our proposed methods.

\subsubsection{Ablation Study}
\begin{table}
  \centering
  \caption{\textbf{Ablation study on the street-view loss} ($\downarrow:$ lower is better; $\uparrow:$ higher is better; \textbf{bold} indicates the best).}
    \setlength{\tabcolsep}{2.0mm}{
    \begin{tabular}{lccc}
    \toprule[1pt]
    \multicolumn{1}{c}{$\text{Loss}$} & \multicolumn{1}{c}{$\text{MAE}\downarrow$} & \multicolumn{1}{c}{$\text{RMSE}\downarrow$} & \multicolumn{1}{c}{$\text{SSIM}\uparrow$}\\
    \midrule[.5pt]
    No Street Loss & $1.1823$  & $ 1.8550 $ & $0.7246$ \\
    Ranking & $1.1783$  & $ 1.8521$ & $\textbf{0.7281}$ \\
    Sky-masked & $1.1808$  & $1.8528$ & $ 0.7229$ \\
    Ranking + Sky-masked & $\textbf{1.1721}$  & $\textbf{1.8407}$ & $0.7196$ \\
    \bottomrule[1pt]
    \end{tabular}}
  \label{tab:ablation-loss}
\end{table}

\begin{figure}
    \centering
    \includegraphics[width=1.0\linewidth]{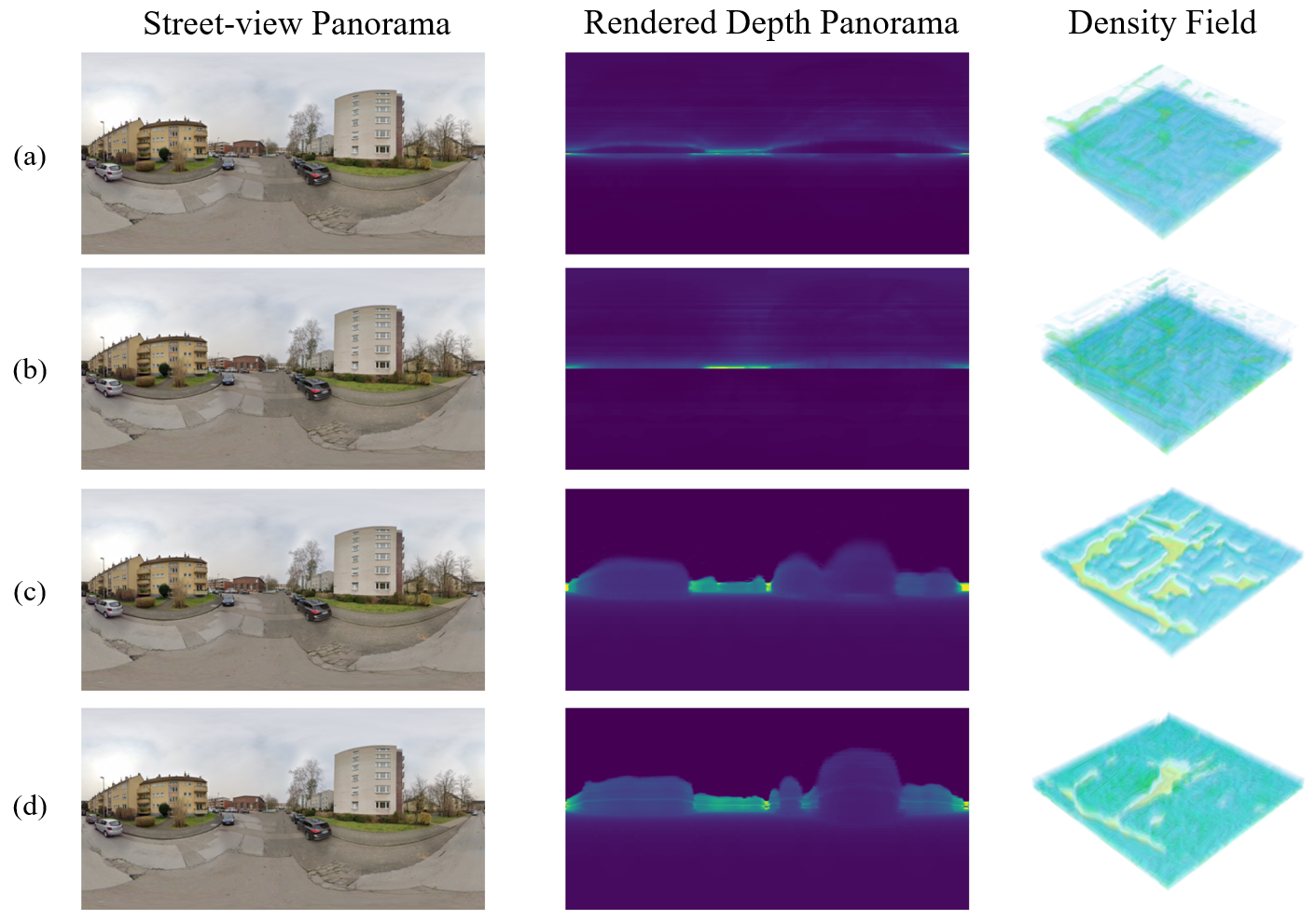}
    \caption{\textbf{Ablation study on the rendered depth panorama and the density field}: (a): density field only (b): density field + fusion (c): density field + street-view loss (d): density field + street-view loss + fusion.}
    \label{fig:ablation-density}
\end{figure}
In the ablation study, we first show the impact of each loss term of the street-view loss, as seen in Table~\ref{tab:ablation-loss}. 
When incorporating either the ranking loss or the sky-masked loss, the performance improves compared to not using any street loss. The ranking loss, when used alone, provides the best SSIM score, indicating its capability of transferring the structural information of the street-view data to the height prediction. Besides, using the sky-masked loss produces similar numerical results.
When combining the ranking and sky-masked losses, our model produces the best height estimation results. However, the SSIM score drops to the lowest. This suggests that while the combined loss terms help refine height predictions, a trade-off exists between achieving higher accuracy and maintaining structural integrity in the predicted results.

Despite the limited performance gains from the street-view loss, from Figure~\ref{fig:ablation-density}, we can find that this loss has a significant impact on the learning of density fields and the rendering of depth panoramas.
First, the model fails to learn a density field with distinct geometry when supervised by only ground truth height maps, as seen in (a) and (b), in which the density fields appear to consist of several overlaid 2D layers, with no meaningful ground objects appearing in the rendered depth panoramas as well.
Next, when including street-view supervision, as seen in (c) and (d), the model manages to learn a geometrically meaningful density field. Even considering the limited coverage of street-view images relative to aerial images, our model is guided to generate a geometrically distinct density field across the entire scene. Additionally, the joint use of feature fusion and street-view loss further enhances the geometry in the rendered depth panorama, which manifests as sharper boundaries in the rendered depth panorama. 

In summary, our findings suggest that the street-view loss can guide the direction of convergence of the network, allowing it to learn density fields with distinct geometric features, thus yielding more accurate height predictions.

\section{Conclusion}\label{sec5}
In this paper, we explore a new problem - enhancing height estimation from monocular aerial imagery using street-view images. Our insight is that the observable vertical structural information in street-view images can help mitigate the scale ambiguity caused by the inherent ill-posed problem in estimating height from a single aerial image. 
Our proposed method for this problem leverages the density field as the intermediary to bridge aerial and street-view images, demonstrating the potential of exploiting additional data sources with rich contextual information for more precise and robust height estimation tasks. 
Moreover, our extension of the GeoNRW dataset facilitates tasks that require co-located overhead and street-view images, such as height estimation and cross-view image synthesis. 

In our future work, we aim to develop more effective cross-modal image fusion methods to further exploit the potential of street-view data and enhance the accuracy of 3D reconstruction tasks that require the use of monocular remote sensing imagery.

{\small
\bibliographystyle{ieee_fullname}
\bibliography{egbib}

\begin{thebibliography}{10}\itemsep=-1pt

\bibitem{9406194}
Gerald Baier, Antonin Deschemps, Michael Schmitt, and Naoto Yokoya.
\newblock Synthesizing optical and sar imagery from land cover maps and
  auxiliary raster data.
\newblock {\em IEEE Transactions on Geoscience and Remote Sensing}, 60:1--12,
  2022.

\bibitem{cao2022monoscene}
Anh-Quan Cao and Raoul de Charette.
\newblock Monoscene: Monocular 3d semantic scene completion.
\newblock In {\em Proceedings of the IEEE/CVF Conference on Computer Vision and
  Pattern Recognition}, pages 3991--4001, 2022.

\bibitem{chen2016single}
Weifeng Chen, Zhao Fu, Dawei Yang, and Jia Deng.
\newblock Single-image depth perception in the wild.
\newblock {\em Advances in neural information processing systems}, 29, 2016.

\bibitem{dosovitskiy2020image}
Alexey Dosovitskiy, Lucas Beyer, Alexander Kolesnikov, Dirk Weissenborn,
  Xiaohua Zhai, Thomas Unterthiner, Mostafa Dehghani, Matthias Minderer, Georg
  Heigold, Sylvain Gelly, et~al.
\newblock An image is worth 16x16 words: Transformers for image recognition at
  scale.
\newblock {\em arXiv preprint arXiv:2010.11929}, 2020.

\bibitem{eigen2014depth}
David Eigen, Christian Puhrsch, and Rob Fergus.
\newblock Depth map prediction from a single image using a multi-scale deep
  network.
\newblock {\em Advances in neural information processing systems}, 27, 2014.

\bibitem{flynn2016deepstereo}
John Flynn, Ivan Neulander, James Philbin, and Noah Snavely.
\newblock Deepstereo: Learning to predict new views from the world's imagery.
\newblock In {\em Proceedings of the IEEE conference on computer vision and
  pattern recognition}, pages 5515--5524, 2016.

\bibitem{fu2018deep}
Huan Fu, Mingming Gong, Chaohui Wang, Kayhan Batmanghelich, and Dacheng Tao.
\newblock Deep ordinal regression network for monocular depth estimation.
\newblock In {\em Proceedings of the IEEE conference on computer vision and
  pattern recognition}, pages 2002--2011, 2018.

\bibitem{8306501}
Pedram Ghamisi and Naoto Yokoya.
\newblock Img2dsm: Height simulation from single imagery using conditional
  generative adversarial net.
\newblock {\em IEEE Geoscience and Remote Sensing Letters}, 15(5):794--798,
  2018.

\bibitem{goodfellow2020generative}
Ian Goodfellow, Jean Pouget-Abadie, Mehdi Mirza, Bing Xu, David Warde-Farley,
  Sherjil Ozair, Aaron Courville, and Yoshua Bengio.
\newblock Generative adversarial networks.
\newblock {\em Communications of the ACM}, 63(11):139--144, 2020.

\bibitem{googlestreetview}
Google.
\newblock Google map street view static api.
\newblock
  \url{https://developers.google.com/maps/documentation/streetview/overview},
  August 2023.

\bibitem{googlemaps}
Google.
\newblock Google maps.
\newblock \url{https://www.google.com/maps}, August 2023.

\bibitem{he2016deep}
Kaiming He, Xiangyu Zhang, Shaoqing Ren, and Jian Sun.
\newblock Deep residual learning for image recognition.
\newblock In {\em Proceedings of the IEEE conference on computer vision and
  pattern recognition}, pages 770--778, 2016.

\bibitem{isola2017image}
Phillip Isola, Jun-Yan Zhu, Tinghui Zhou, and Alexei~A Efros.
\newblock Image-to-image translation with conditional adversarial networks.
\newblock In {\em Proceedings of the IEEE conference on computer vision and
  pattern recognition}, pages 1125--1134, 2017.

\bibitem{KAKU2019417}
Kazuya Kaku.
\newblock Satellite remote sensing for disaster management support: A holistic
  and staged approach based on case studies in sentinel asia.
\newblock {\em International Journal of Disaster Risk Reduction}, 33:417--432,
  2019.

\bibitem{rs13122417}
Savvas Karatsiolis, Andreas Kamilaris, and Ian Cole.
\newblock Img2ndsm: Height estimation from single airborne rgb images with deep
  learning.
\newblock {\em Remote Sensing}, 13(12), 2021.

\bibitem{kingma2014adam}
Diederik~P Kingma and Jimmy Ba.
\newblock Adam: A method for stochastic optimization.
\newblock {\em arXiv preprint arXiv:1412.6980}, 2014.

\bibitem{lemmens1988survey}
MJPM Lemmens.
\newblock A survey on stereo matching techniques.
\newblock {\em International Archives of Photogrammetry and Remote Sensing},
  27(B8):11--23, 1988.

\bibitem{liasis2016satellite}
Gregoris Liasis and Stavros Stavrou.
\newblock Satellite images analysis for shadow detection and building height
  estimation.
\newblock {\em ISPRS Journal of Photogrammetry and Remote Sensing},
  119:437--450, 2016.

\bibitem{lienen2021monocular}
Julian Lienen, Eyke Hullermeier, Ralph Ewerth, and Nils Nommensen.
\newblock Monocular depth estimation via listwise ranking using the
  plackett-luce model.
\newblock In {\em Proceedings of the IEEE/CVF Conference on Computer Vision and
  Pattern Recognition}, pages 14595--14604, 2021.

\bibitem{rs12172719}
Chao-Jung Liu, Vladimir~A. Krylov, Paul Kane, Geraldine Kavanagh, and Rozenn
  Dahyot.
\newblock Im2elevation: Building height estimation from single-view aerial
  imagery.
\newblock {\em Remote Sensing}, 12(17), 2020.

\bibitem{long2015fully}
Jonathan Long, Evan Shelhamer, and Trevor Darrell.
\newblock Fully convolutional networks for semantic segmentation.
\newblock In {\em Proceedings of the IEEE conference on computer vision and
  pattern recognition}, pages 3431--3440, 2015.

\bibitem{Lu_2020_CVPR}
Xiaohu Lu, Zuoyue Li, Zhaopeng Cui, Martin~R. Oswald, Marc Pollefeys, and
  Rongjun Qin.
\newblock Geometry-aware satellite-to-ground image synthesis for urban areas.
\newblock In {\em Proceedings of the IEEE/CVF Conference on Computer Vision and
  Pattern Recognition (CVPR)}, June 2020.

\bibitem{bingmaps}
Microsoft.
\newblock Bing maps.
\newblock \url{https://www.bing.com/maps}, August 2023.

\bibitem{mildenhall2021nerf}
Ben Mildenhall, Pratul~P Srinivasan, Matthew Tancik, Jonathan~T Barron, Ravi
  Ramamoorthi, and Ren Ng.
\newblock Nerf: Representing scenes as neural radiance fields for view
  synthesis.
\newblock {\em Communications of the ACM}, 65(1):99--106, 2021.

\bibitem{mirza2014conditional}
Mehdi Mirza and Simon Osindero.
\newblock Conditional generative adversarial nets.
\newblock {\em arXiv preprint arXiv:1411.1784}, 2014.

\bibitem{mou2018im2height}
Lichao Mou and Xiao~Xiang Zhu.
\newblock Im2height: Height estimation from single monocular imagery via fully
  residual convolutional-deconvolutional network.
\newblock {\em arXiv preprint arXiv:1802.10249}, 2018.

\bibitem{dsm2dtm}
Rajat Naman~Jain.
\newblock {dsm2dtm}.
\newblock \url{https://github.com/seedlit/dsm2dtm}, August 2023.

\bibitem{park2018high}
Kihong Park, Seungryong Kim, and Kwanghoon Sohn.
\newblock High-precision depth estimation with the 3d lidar and stereo fusion.
\newblock In {\em 2018 IEEE International Conference on Robotics and Automation
  (ICRA)}, pages 2156--2163. IEEE, 2018.

\bibitem{PARK201976}
Yujin Park and Jean-Michel Guldmann.
\newblock Creating 3d city models with building footprints and lidar point
  cloud classification: A machine learning approach.
\newblock {\em Computers, Environment and Urban Systems}, 75:76--89, 2019.

\bibitem{NEURIPS2019_9015}
Adam Paszke, Sam Gross, Francisco Massa, Adam Lerer, James Bradbury, Gregory
  Chanan, Trevor Killeen, Zeming Lin, Natalia Gimelshein, Luca Antiga, Alban
  Desmaison, Andreas Kopf, Edward Yang, Zachary DeVito, Martin Raison, Alykhan
  Tejani, Sasank Chilamkurthy, Benoit Steiner, Lu Fang, Junjie Bai, and Soumith
  Chintala.
\newblock Pytorch: An imperative style, high-performance deep learning library.
\newblock In {\em Advances in Neural Information Processing Systems 32}, pages
  8024--8035. Curran Associates, Inc., 2019.

\bibitem{qian2023sat2density}
Ming Qian, Jincheng Xiong, Gui-Song Xia, and Nan Xue.
\newblock Sat2density: Faithful density learning from satellite-ground image
  pairs.
\newblock {\em arXiv preprint arXiv:2303.14672}, 2023.

\bibitem{ranftl2021vision}
Ren{\'e} Ranftl, Alexey Bochkovskiy, and Vladlen Koltun.
\newblock Vision transformers for dense prediction.
\newblock In {\em Proceedings of the IEEE/CVF international conference on
  computer vision}, pages 12179--12188, 2021.

\bibitem{ranftl2020towards}
Ren{\'e} Ranftl, Katrin Lasinger, David Hafner, Konrad Schindler, and Vladlen
  Koltun.
\newblock Towards robust monocular depth estimation: Mixing datasets for
  zero-shot cross-dataset transfer.
\newblock {\em IEEE transactions on pattern analysis and machine intelligence},
  44(3):1623--1637, 2020.

\bibitem{Regmi_2018_CVPR}
Krishna Regmi and Ali Borji.
\newblock Cross-view image synthesis using conditional gans.
\newblock In {\em Proceedings of the IEEE Conference on Computer Vision and
  Pattern Recognition (CVPR)}, June 2018.

\bibitem{ronneberger2015u}
Olaf Ronneberger, Philipp Fischer, and Thomas Brox.
\newblock U-net: Convolutional networks for biomedical image segmentation.
\newblock In {\em Medical Image Computing and Computer-Assisted
  Intervention--MICCAI 2015: 18th International Conference, Munich, Germany,
  October 5-9, 2015, Proceedings, Part III 18}, pages 234--241. Springer, 2015.

\bibitem{schmidhuber2015deep}
J{\"u}rgen Schmidhuber.
\newblock Deep learning in neural networks: An overview.
\newblock {\em Neural networks}, 61:85--117, 2015.

\bibitem{shi2022geometry}
Yujiao Shi, Dylan Campbell, Xin Yu, and Hongdong Li.
\newblock Geometry-guided street-view panorama synthesis from satellite
  imagery.
\newblock {\em IEEE Transactions on Pattern Analysis and Machine Intelligence},
  44(12):10009--10022, 2022.

\bibitem{ten2019biomass}
Jelle ten Harkel, Harm Bartholomeus, and Lammert Kooistra.
\newblock Biomass and crop height estimation of different crops using uav-based
  lidar.
\newblock {\em Remote Sensing}, 12(1):17, 2019.

\bibitem{vaswani2017attention}
Ashish Vaswani, Noam Shazeer, Niki Parmar, Jakob Uszkoreit, Llion Jones,
  Aidan~N Gomez, {\L}ukasz Kaiser, and Illia Polosukhin.
\newblock Attention is all you need.
\newblock {\em Advances in neural information processing systems}, 30, 2017.

\bibitem{wang2004image}
Zhou Wang, Alan~C Bovik, Hamid~R Sheikh, and Eero~P Simoncelli.
\newblock Image quality assessment: from error visibility to structural
  similarity.
\newblock {\em IEEE transactions on image processing}, 13(4):600--612, 2004.

\bibitem{workman2021augmenting}
Scott Workman and Hunter Blanton.
\newblock Augmenting depth estimation with geospatial context.
\newblock In {\em Proceedings of the IEEE/CVF International Conference on
  Computer Vision}, pages 4562--4571, 2021.

\bibitem{rs13152862}
Yakun Xie, Dejun Feng, Sifan Xiong, Jun Zhu, and Yangge Liu.
\newblock Multi-scene building height estimation method based on shadow in high
  resolution imagery.
\newblock {\em Remote Sensing}, 13(15), 2021.

\bibitem{yoo20203d}
Jin~Hyeok Yoo, Yecheol Kim, Jisong Kim, and Jun~Won Choi.
\newblock 3d-cvf: Generating joint camera and lidar features using cross-view
  spatial feature fusion for 3d object detection.
\newblock In {\em Computer Vision--ECCV 2020: 16th European Conference,
  Glasgow, UK, August 23--28, 2020, Proceedings, Part XXVII 16}, pages
  720--736. Springer, 2020.

\bibitem{zhang2022bending}
Jiaming Zhang, Kailun Yang, Chaoxiang Ma, Simon Rei{\ss}, Kunyu Peng, and
  Rainer Stiefelhagen.
\newblock Bending reality: Distortion-aware transformers for adapting to
  panoramic semantic segmentation.
\newblock In {\em 2022 IEEE/CVF Conference on Computer Vision and Pattern
  Recognition (CVPR)}, pages 16917--16927, 2022.

\bibitem{zhu2017toward}
Jun-Yan Zhu, Richard Zhang, Deepak Pathak, Trevor Darrell, Alexei~A Efros,
  Oliver Wang, and Eli Shechtman.
\newblock Toward multimodal image-to-image translation.
\newblock {\em Advances in neural information processing systems}, 30, 2017.

\end{thebibliography}
}

\end{document}